%% file: root.tex
\documentclass[letterpaper, 10 pt, conference]{ieeeconf}  % Comment this line out if you need a4paper

\IEEEoverridecommandlockouts                              % This command is only needed if 
\usepackage{epsfig} % for postscript graphics files
\usepackage{times} % assumes a new font selection scheme installed
\usepackage{amsmath} % assumes amsmath package installed
\usepackage{amssymb}  % assumes amsmath package installed
\usepackage{gensymb}
\usepackage{booktabs}
\usepackage{bm}
\usepackage{url}
\usepackage{nccmath}
\usepackage{siunitx}
\usepackage{wrapfig}
\usepackage{framed}
\usepackage{multirow}
\usepackage[font=small, labelfont=bf]{caption}
\usepackage{arydshln}
\usepackage{makecell}
\usepackage{xcolor}
\usepackage{hyperref}
\usepackage{array}
\usepackage{ragged2e}
\usepackage{tabularray}

\newcommand{\rectangle}{\fboxsep0pt\fbox{\rule{1em}{0pt}\rule{0pt}{1ex}}}

\title{\LARGE \bf
Robust and Resilient Soft Robotic Object Insertion with 
Compliance-Enabled Contact Formation and Failure Recovery
}

\author{Mimo Shirasaka$^{1*}$, Cristian C. Beltran-Hernandez$^{1}$, Masashi Hamaya$^{1}$, Yoshitaka Ushiku$^{1}$% <-this % stops a space
\thanks{*This work has been done during an internship. }% <-this % stops a space
\thanks{This work was supported by JST, Moonshot R\&D, Grant Number JPMJMS2236 and JST, PRESTO, Grant Number JPMJPR24T9, Japan.}
\thanks{We used ChatGPT and Claude to proofread all sections of our manuscript.}% <-this % stops a space
\thanks{$^{1}$ OMRON SINIC X Corporation, Tokyo, Japan
        {\tt\small masashi.hamaya@sinicx.com}}%
}

\begin{document}

\maketitle
\thispagestyle{empty}
\pagestyle{empty}

% \makeatletter
% \let\@oldmaketitle\@maketitle%
% \renewcommand{\@maketitle}{\@oldmaketitle%
%     \centering
%     \vspace*{1mm}
%     \includegraphics[width=\textwidth]{figures/system_overview_MS1023 (1).png}
%     \captionof{figure}{Overview of the proposed framework. }
%     \label{fig:overview}
%     \vspace*{-3mm}
%     }
% \maketitle
% \setcounter{figure}{1}

\input{sections/0-abstract}

\input{sections/1-introduction}
\input{sections/2-related}
\input{sections/3-methodology}

\input{sections/4-exp_setup}

\section{Discussion}
\input{sections/discussion}

\section{CONCLUSIONS}  
This study integrates VLM-based compliance-enabled failure recovery into a compliance-enabled contact-formation framework with a soft wrist for robust and resilient peg-in-hole insertion. The simulation and real robot experiments demonstrated that our method effectively recovered from failures and improved success rates under uncertain and unseen conditions.

\bibliographystyle{IEEEtran}
\bibliography{reference}

\end{document}

%% file: sections/0-abstract.tex
%%%%%%%%%%%%%%%%%%%%%%%%%%%%%%%%%%%%%%%%%%%%%%%%%%%%%%%%%%%%%%%%%%%%%%%%%%%%%%%%

\begin{abstract}
Object insertion tasks are prone to failure under pose uncertainty and environmental variation, often requiring manual fine-tuning or controller retraining.
We present a novel approach for robust and resilient object insertion using a passively compliant soft wrist that enables safe contact absorption through large deformations, without high-frequency control or force sensing. 
Our method structures insertion as compliance-enabled contact formations, sequential contact states that progressively constrain degrees of freedom, and integrates automated failure recovery strategies.
Our key insight is that wrist compliance permits safe, repeated recovery attempts; hence, we refer to it as compliance-enabled failure recovery. We employ a pre-trained vision-language model (VLM) that assesses each skill execution from terminal poses and images, identifies failure modes, and proposes recovery actions by selecting skills and updating goals. In simulation, our method achieved an 83\% success rate, recovering from failures induced by randomized conditions, including grasp misalignments up to $5^\circ$, hole-pose errors up to 20\,mm, fivefold increases in friction, and unseen square/rectangular pegs, and we further validated the approach on a real robot.
Project page is available at
\href{https://omron-sinicx.github.io/compliance-enabled-failure-recovery/}
{\nolinkurl{https://omron-sinicx.github.io/compliance-enabled-failure-recovery/}}.
\end{abstract}

%% file: sections/1-introduction.tex
%%%%%%%%%%%%%%%%%%%%%%%%%%%%%%%%%%%%%%%%%%%%%%%%%%%%%%%%%%%%%%%%%%%%%%%%%%%%%%%%
\section{INTRODUCTION}
Peg-in-hole tasks have been widely studied but remain challenging due to their contact-rich nature and tight tolerances~\cite{xu2019compare}.
A central difficulty lies in handling uncertainties in part grasping and hole pose.
Conventional methods address these uncertainties through precise pose estimation and rigid fixturing, enabling repetitive pick-and-place operations~\cite{otaIROS2024}, but such approaches demand substantial engineering effort.
This study aims to realize an insertion strategy that explicitly accounts for uncertainty.

Compliance-enabled strategies mitigate these challenges by tolerating positional variation.
Passive compliance~\cite{Morgan2023TowardsGeneralizedRobotAsembly, nishimura2017peg, ZhangRAL2023} enables stable contact through inherent mechanical deformation, while active compliance~\cite{beltran2020variable} relies on precise force sensing and real-time position control to adaptively adjust contact forces and positions.
We build on soft robots, which inherently provide passive compliance and accommodate positional uncertainty through large elastic deformations without high-frequency control or precise force sensing.

While soft robots naturally absorb uncertainty, a key question remains: how can we develop robust and resilient control strategies that leverage their compliance?
Here, robustness denotes reliable execution under uncertainty, and resilience denotes recovery from failures.

\begin{figure}[t]
    \centering
    \includegraphics[width=\linewidth]{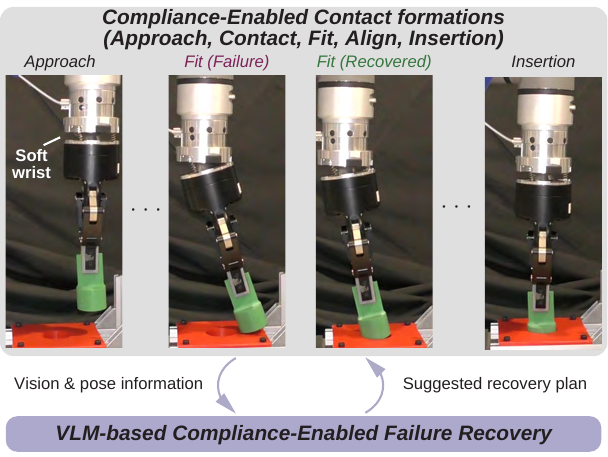}
    \caption{We propose compliance-enabled contact formations and failure recovery with a soft wrist to achieve more robust and resilient object insertion.}
    \label{fig:proposed}
\end{figure}

To enhance robustness, we exploit contact formations~\cite{skubic1996identifying}, discrete contact states that constrain the parts' relative motion during assembly.
Compliance plays a critical role by enabling stable contact and smooth contact-state transitions, a concept termed \emph{compliance-enabled contact formation}~\cite{Morgan2023TowardsGeneralizedRobotAsembly}.
In peg-in-hole insertion, these formations comprise phases such as approach, contact, fit, alignment, and insertion, each restricting specific degrees of freedom.
Following these formations sequentially improves task robustness.
Prior work demonstrates the effectiveness of compliance-enabled contact formation~\cite{Morgan2023TowardsGeneralizedRobotAsembly, johannsmeier2019framework, vuong2021learning}; however, failures still occur under severe uncertainties and environmental variations (e.g., friction, part geometry, material properties).

To achieve both robustness and resilience, we integrate a failure recovery strategy into compliance-enabled contact formation.
Compliance further facilitates recovery: even if an attempt fails, retries remain safe because compliant interactions limit contact forces, hence we refer to it as \emph{compliance-enabled failure recovery}.
A major challenge is accurately identifying failures and selecting appropriate recovery actions.
Previous soft-robot recovery methods required training failure detectors~\cite{gilday2023predictive}.
Inspired by recent advances with Large Language Models (LLMs)~\cite{sinhareal} and Vision-Language Models (VLMs)~\cite{dai2025racer}, we employ a VLM to enable compliance-enabled failure recovery without training (Fig.~\ref{fig:proposed}).
The vision-language modality analyzes and describes failure cases and proposes recovery plans.
Concretely, we predefine goal-conditioned skills for each contact formation and execute them sequentially.
A pre-trained VLM (e.g., GPT-4o) evaluates success or failure and explains the outcome; upon failure, it suggests a recovery plan by selecting a sequence of skills and updating goal parameters.

We validate the approach in simulation and on a real robot using a passive compliant soft wrist~\cite{von2020compact} mounted on a collaborative arm.
The method recovers from failures via the VLM in previously unseen settings, including uncertainties in grasping or hole pose and variations in contact friction or peg geometry.

{\bf Contribution:}
We develop a compliance-enabled contact-formation and failure-recovery framework for peg-in-hole insertion, combining a soft wrist with VLM-based reasoning.

%% file: sections/2-related.tex
\section{RELATED WORK}
\subsection{Robust and resilient soft robots}
Soft robots conform to object geometry and absorb disturbances, and have therefore been used in applications such as grippers and assistive robots~\cite{agarwal2016stretchable, homberg2019robust, li2022soft}. Prior work on resilience in soft robotics has focused on damage tolerance and self-healing systems~\cite{bilodeau2017self}. While substantial effort has addressed robustness and resilience from a mechanical perspective, this work targets robust and resilient \emph{control} for peg-in-hole insertion by leveraging soft robots' inherent compliance. The following subsections discuss compliance-enabled contact formation and failure recovery, which together enhance robustness and resilience.

\subsection{Compliance-enabled contact formation}
Contact formations enhance robust contact-rich manipulation. Model-based approaches employ velocity control to achieve the target constraint forces~\cite{Morgan2023TowardsGeneralizedRobotAsembly} for each contact formation. 
Learning-based techniques update impedance control parameters using Bayesian optimization~\cite{johannsmeier2019framework} and reinforcement learning~\cite{vuong2021learning}. 
Additionally, learning from demonstrations has been used to extract skills for contact-rich manipulation tasks~\cite{le2021learning, okada2023learning}.
Hamaya et al. segmented a peg-in-hole task into several contact formations and employed reinforcement learning~\cite{HamayaICRA2020}.
Wang et al. proposed contact configuration identification for hole search using action primitives~\cite{wang2023pomdp}.
Zhang et al. introduced a reinforcement learning method composed of discrete and continuous action spaces to select insertion primitives and low-level control parameters~\cite{zhang2022learning}.
Saito et al. proposed training action primitives based on the contact state transitions for in-hand manipulation~\cite{saito2024apricot}.
Kim et al. proposed using tactile sensors to localize the contact line between the peg and hole~\cite{kim2022active}.

Although the methods described above demonstrated robustness in contact-rich manipulations, explicit failure recovery has not been addressed.
A recent study by Wu et al. designed a finite-state machine for recovery, which switches skills automatically based on high-frequency force sensor readings and controller~\cite {wu20241}.  
Our study proposes using a soft wrist and VLM to suggest recovery plans automatically in contact formation-based manipulation.

\subsection{Learning-based error detection and recovery}
Numerous researchers have investigated error detection and recovery strategies. 
Anomaly detection methods benefit robots by alerting them to dangerous or unseen states.
Romeres et al. used a Gaussian process-based  
anomaly detection method when the measured force deviates significantly from the expected range~\cite{romeres2019anomaly}.
Park et al. proposed a vision-audio-force-based error detection method composed of an LSTM-based autoencoder~\cite{park2018multimodal}.

Learning-based approaches enable robots to obtain recovery strategies automatically.   
Chang et al.~employed a Petri net for failure recovery; when the robot fails and visits an unknown state, it asks for human demonstrations to correct the error~\cite{chang2013robot}.
Hegemann et al. proposed a method for learning skill graphs based on human labeling and employed multimodal sensors for error detection in grasping and mobile manipulation tasks~\cite{hegemann2022learning}.
Thananjeyan et al. proposed a recovery reinforcement learning method that executed a recovery policy when the agent entered unsafe areas~\cite{thananjeyan2021recovery}.
Vats et al.~proposed an efficient learning method that incrementally discovers failure cases and subsequently learns recovery skills~\cite{vats2023efficient}.
Saxena et al.~developed a vision-based failure predictor and corresponding recovery maneuvers for aerial robots~\cite{saxena2017learning}. 

Recent strategies based on LLM or VLM have demonstrated self-recovery in various tasks.
Dai et al. proposed combining a VLM with a language-guided visuomotor policy, where the VLM detects errors and suggests recovery strategies using rich language instructions~\cite{dai2025racer}.
Cornelio et al. integrated ontologies, logical rules, and LLM-based planners for online error detection and recovery in simulated kitchen environments~\cite{cornelio2024recover}.
Shinha et al. introduced an LLM-based fast-slow reasoning method for aerial robots, which employs fast anomaly detection in the LLM's embedding space and slow autoregressive generation to determine the criticality of anomalies~\cite{sinhareal}. 
Liu et al. developed a framework for robot failure explanation and correction in which an LLM verifies the success of each subgoal based on vision and audio summaries~\cite{liu2023reflect}.  
Shirasaka et al. proposed a self-recovery prompting pipeline for promptable robot systems, which refines prompts based on past experiences and active communication with the operator to recover from failures~\cite{shirasaka2024self}.
Wang et al. combined task and motion planning with an LLM to automatically generate recovery symbolic plans~\cite{wang2024llm}.

While these methods recover from failures across diverse tasks, they primarily target rigid-robot pick-and-place and mobile navigation. A notable exception in soft robotics proposes a tactile-based predictive network that classifies grasp success or failure~\cite{gilday2023predictive}, but it requires data collection and model training. In contrast, we address peg-in-hole insertion by combining a pretrained VLM-based failure-recovery strategy with compliance-enabled contact formation using a soft wrist.

%% file: sections/3-methodology.tex
\begin{figure*}[t]
    \centering
    \includegraphics[width=\linewidth]{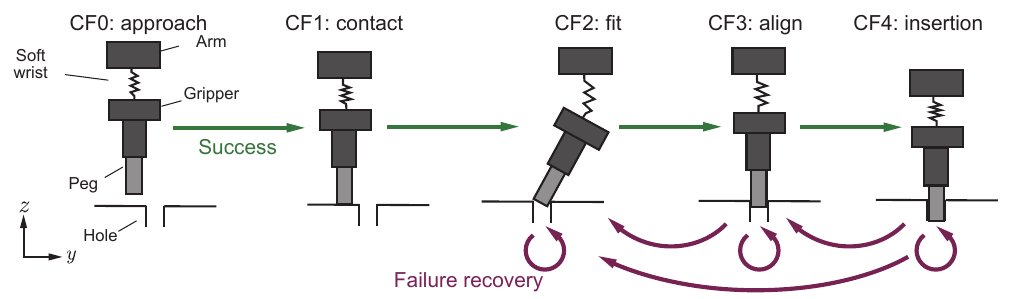}
    \caption{Compliance-enabled contact formation: We employ the contact formations for peg-in-hole tasks by following the prior studies~\cite{johannsmeier2019framework, HamayaICRA2020}. It consists of approach, contact, fit, align, and insertion contact formations. We consider applying a failure recovery strategy in fit, align, and insertion contact formations.}
    \label{fig:cf}
\end{figure*}

\section{FRAMEWORK OVERVIEW}
We aim to robustly and resiliently complete peg-in-hole tasks, even when faced with uncertainties in grasp or hole pose estimations and environmental variations such as unseen surface friction and peg shapes.
This study uses a rigid robot arm and a two-finger gripper with a soft wrist~\cite{von2020compact}.
This soft wrist has three springs and allows large 6-DoF deformations.
An RGB camera is used to observe the peg and the hole.  

Following the prior studies~\cite{johannsmeier2019framework, HamayaICRA2020}, we prepare manipulation skills and the goal information of the rigid robot arm's poses $p_g$ and images $I_g$ for the predefined contact formations. 
The terminal information, which consists of the robot arm pose $p_e$ and the image $I_e$, is received at the timestep when the skill executions are completed or halted.  

Each $n \in N$ skill executes position control sequentially given $p^n_g$.
A VLM evaluates the success or failure of skill $n$ based on the goal and terminal information. 

If it fails, the VLM suggests a recovery plan that involves reorganizing the skill sequences and updating the goal parameters as necessary.
Secs.~\ref{sec:skill} and \ref{sec:vlm} describe more detailed information on the contact formations for peg-in-hole tasks and failure recovery strategies, respectively.

\section{MANIPULATION SKILLS IN COMPLIANCE-ENABLED CONTACT FORMATION}
\label{sec:skill}
This section introduces compliance-enabled contact formation for peg-in-hole tasks. Following the previous studies~\cite{johannsmeier2019framework, HamayaICRA2020}, we assume a planar (2D) peg insertion and a fixed height of the hole surface for simplicity; the arm pose dimensions are $y$, $z$, and roll $\theta$.
% and the force is $f_y$, $f_z$, $\tau_x$.
As shown in Fig.~\ref{fig:cf}, peg-in-hole tasks have five contact formations (CF): approach, contact, fit, align, and insertion.
Only the dimensions of interest are evaluated and updated in each CF.

{\bf CF0~(Approach)}: The robot moves close to the board. No contact occurs. The goal pose is given as:
\begin{equation*}
    p_g^{n=0} = [y, z, \theta].
\end{equation*}

{\bf CF1~(Contact)}: The robot goes straight down until it touches the board's surface. Only the $z$-axis goal pose is provided or updated. 
\begin{equation*}
    p_g^{n=1} = [-, z, -],
\end{equation*}
where $-$ represents the goal pose of the previous skill.

{\bf CF2~(Fit)}: The robot slides the peg horizontally until the peg's tip touches the hole's edge:
\begin{equation*}
    p_g^{n=2} = [y, -, -].
\end{equation*}
In CF2, as the soft wrist allows stable contact for the $z$-axis, we can only consider the $y$-axis goal pose.

{\bf CF3 (Align)}: The robot rotates the peg to align it vertically:
\begin{equation*}
    p_g^{n=3} = [y, -, -].
\end{equation*}
Due to its compliance, the soft wrist can automatically rotate the peg with only $y$-axis translations.

{\bf CF4 (Insertion)}: The robot moves the peg down to a certain depth of the hole.
We consider only the $z$-axis motion: 
\begin{equation*}
    p_g^{n=4} = [-, z, -].
\end{equation*}

We sequentially execute the skills of CF0 to CF4. When the CF4 is completed, the overall task is completed. 
Due to the fixed height of the hole surface, CF0 and CF1 are assumed to succeed under our setup.
When the robot fails in a CF, the recovery plan suggests the subsequent skill sequences starting from CF2 to CF4.

\section{COMPLIANCE-ENABLED FAILURE RECOVERY WITH VLM}
\label{sec:vlm}
\begin{figure}[t]
    \centering \includegraphics[width=\linewidth]{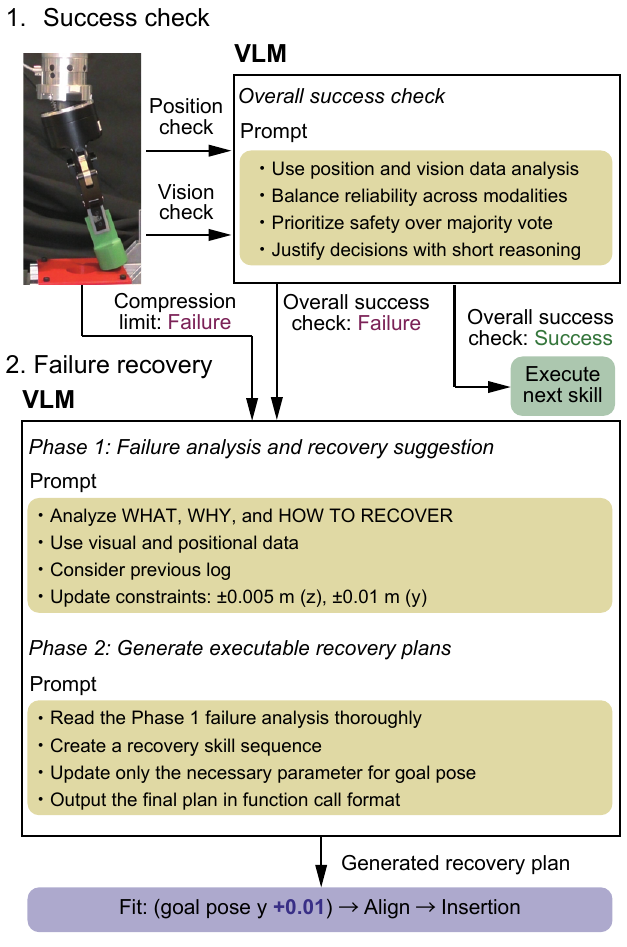}
    \caption{Overview of our failure recovery strategy: 1) Success check, where a VLM evaluates the skill execution. 2) Failure recovery planning, where the VLM analyzes the failure and generates recovery plans.}
    \label{fig:failure_recovery}
\end{figure}

This section introduces our method for compliance-enabled failure recovery, which consists of a success check and a failure recovery process.
The overview is shown in Fig.~\ref{fig:failure_recovery}.

First, a VLM judges success or failure based on the terminal robot arm pose $p_e$ and the image $I_e$.
If the VLM judges that the skill execution of the contact formation fails, the VLM-based failure recovery is executed. 
The VLM analyzes the reasons for failure, suggests recovery skill sequences, and updates goal parameters as necessary. 

\begin{table}[t]
  \centering
  \caption{OpenCV Success Criteria. Here, $p_{ey}$ and $p_{gy}$ are the y-components of $p_e$ and $p_g$, respectively.}
  \label{tab:opencv_criteria}
  \begin{tabular}{ccc}
     Fit \\
    \toprule
    $P_{center}$ & $P_{edge}$ & Result Message \\
    \midrule
    $\leq H_{center}$ & $ > H_{center}$ & success \\
    $\leq H_{center}$ & $\leq H_{center}$ & $p_{gy}$ was too small. \\
    otherwise & -- & $p_{gy}$ was too big. \\
    \bottomrule
    \\
    Align \\
    \toprule
    Peg relative pose & Tilt Angle & Result Message \\
    \midrule
    On the hole & Within $\pm$5° & 
    \makecell[l]{Align success} \\
    On the hole & $>$ 5° & 
    \makecell[l]{peg is tilted to the left.} \\
    On the hole & $<$ $-5$° & \makecell[l]{peg is tilted to the right.} \\
    Left of the hole & Within $\pm$5° & 
    \makecell[l]{$p_{ey}$ was too small.} \\
    Right of the hole & Within $\pm$5° & 
    \makecell[l]{$p_{ey}$ was too big.} \\
    Left of the hole & $>$ 5° & \makecell[l]{peg is tilted to the left.\\$p_{gy}$ was too small.} \\
    Right of the hole& $>$ 5° & \makecell[l]{$p_{gy}$ was too big.\\also tilted to the left.} \\
    Left of the hole& $<$ $-5$° & \makecell[l]{ $p_{gy}$ was too small.\\also tilted to the right.} \\
    Right of the hole& $<$ $-5$° & \makecell[l]{$p_{gy}$ was too big.\\also tilted to the right.} \\
    \bottomrule
  \end{tabular}
\end{table}

\begin{figure}
    \centering
  \includegraphics[width=\linewidth]{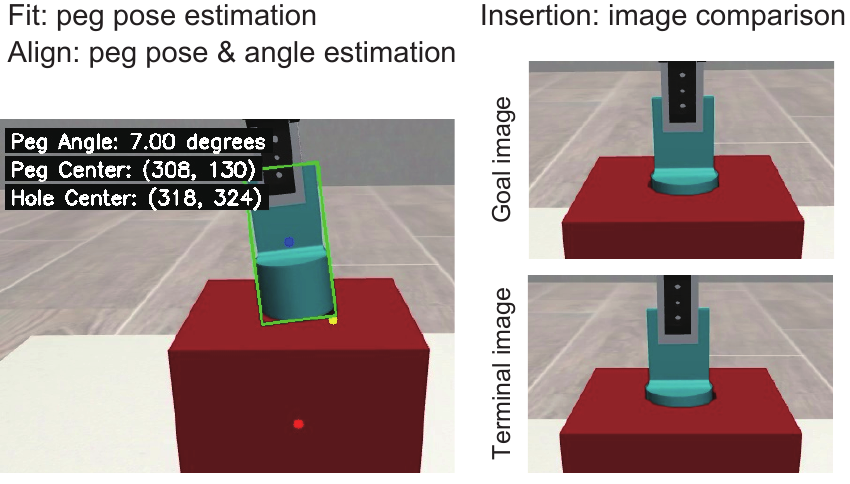}
    \caption{Vision success check: we applied vision-based pose and angle estimation using OpenCV in fit and align, and image comparison that is fed to a VLM in insertion. }
    \label{fig:visual_success_examples}
\end{figure}

\subsection{Success check}
\label{sec:multimodal_verification}

{\bf Vision and pose-based success check:}
% vision success verification
To verify success in the Fit and Align CFs, we apply vision-based analysis using OpenCV. The peg and hole are segmented using HSV color thresholding: the peg is detected as a green (or cyan in simulation) region, and the hole as a red region. We extract the largest contour for each object to estimate its bounding boxes and center coordinates. For pose estimation, we fit a minimum-area rectangle to the peg contour using \texttt{cv2.minAreaRect}, from which the tilt angle of the peg is computed. 
Success is determined based on visual measurements of the peg’s position and orientation relative to the hole. For the Fit subtask, we compare the y-coordinates of the peg’s center $P_{center}$ and its bottom-right edge $P_{edge}$ against the hole’s center $H_{center}$. For the Align subtask, we evaluate (1) the horizontal offset between $P_{center}$ and $H_{center}$, and (2) whether the peg’s tilt is within $\pm$5{\degree} of vertical. The criteria for both subtasks are summarized in Table~\ref{tab:opencv_criteria}.

In the Insertion subtask, the peg becomes visually hidden once it enters the hole, making it difficult for OpenCV to determine whether the operation was successful. Therefore, we provide a goal image and success criteria in the prompt and use a VLM to determine vision-based success. The visual success verification overview is shown in Fig.~\ref{fig:visual_success_examples}.

We also apply a pose-based success check to determine whether the robot reaches the goal pose. The success condition is determined as: $\|p_g - p_e\| \leq p_{th}$, where $p_{th}$ is a threshold for the pose error, set to 0.001 m.

{\bf Compression limit:} 
During skill execution, the compression limit is reached when the soft-wrist springs are compressed along the 
$z$-axis beyond a predefined threshold. When this occurs, the skill terminates immediately, and the failure-recovery procedure is activated.

{\bf Overall success check by VLM: }
% overall analysis with VLM
A VLM performs an overall analysis of execution results for each skill, given the results of the vision- and pose-based success checks. In precision tasks, multiple sensors may provide conflicting answers (i.e., differing decisions regarding the success or failure of each subtask) due to recognition errors. We leverage the VLM’s reasoning ability to integrate multisensory analysis results and make a final decision on subtask success. The VLM returns a success or failure result; if the result indicates failure, it moves to the failure recovery stage.

\subsection{Failure recovery with VLM}
The failure recovery system consists of a two-stage analysis with VLM. In the first phase, a VLM analyzes the failure and suggests a recovery plan, including a recovery motion as a relative translation. Based on this analysis, the second VLM agent generates a recovery plan in an executable format with absolute goal positions reflecting the suggested relative movement in the previous analysis. Below, we provide a detailed explanation of the analysis flow.

{\bf Failure analysis and recovery suggestion:}
To generate a concise recovery plan, an accurate understanding of the situation is essential. Thus, upon detecting a failure, the framework summarizes the situation, focusing on the failure. We establish an analysis-to-suggestion thought pathway by prompting the VLM to consider the situation from three perspectives: 1) What happened?, 2) Why did it occur?, and 3) How can the framework recover from the failure?
The recovery suggestion includes motion instructions in terms of direction and values relative to the current goal parameters. By guiding the VLM to reason through the situation, we enable the recovery framework to generate a well-founded strategy, thereby enhancing both the explainability of the recovery process and the transparency of the framework’s workflow.

{\bf Executable recovery plan generation:}
In the second phase of failure recovery, the VLM translates the suggested recovery strategy into a sequence of executable skills necessary to complete the task (i.e., successfully inserting the peg into the hole). To achieve this, we utilize the VLM’s function-calling capability, prompting it to update the goal position as needed. These updates reflect the recovery adjustments proposed in the previous phase, ensuring a structured and adaptive execution plan.

%% file: sections/4-exp_setup.tex
\section{EXPERIMENTS}
We performed experiments in a simulation and with a real robot. 
The robotic system consisted of a Universal Robot UR5e, a Robotiq Adaptive Gripper 2-Finger 85, and a soft wrist~\cite{von2020compact}. A built-in force/torque sensor of UR5e was used only to check the compression limit of the wrist's springs, and was not used for control. The soft wrist comprises three springs that allow large 6-DoF deformations. For VLM, we used GPT-4o (version 1.64.0).

The goal of the experiments is to answer the following questions: (1) Can our method improve the success rates through recovery?, (2) Can our method recover from failures in uncertain and unseen environments?, and (3) Can our method be used in real-world environments? 
\begin{figure*}[t]
    \centering
    \includegraphics[width=\linewidth]{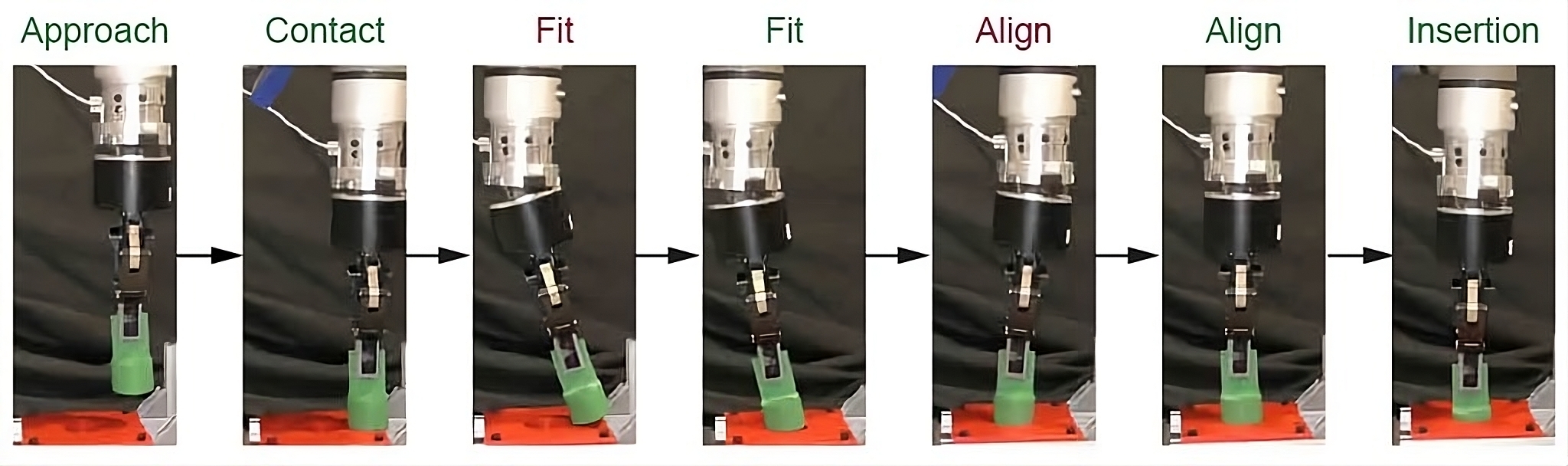}
    \caption{Snapshots of a peg-in-hole trial: The robot failed in the Fit and Align contact formations, but it recovered using our method. The contact formations colored with \textcolor[rgb]{0.066, 0.466, 0.200}{green} and \textcolor[rgb]{0.533, 0.133, 0.333}{purple} above each snapshot describe the successful and failed executions, respectively.}
    \label{fig:snapshots}
\end{figure*}

\subsection{Simulation experiments}
To answer the first and second questions, we performed simulation experiments. 

{\bf Simulator setup:} We used a simulator developed by a previous study~\cite{fuchioka2024robotic}. The simulation is based on the Robosuite framework with the MuJoCo physics engine~\cite{zhu2020robosuite}. 
The previous study developed a simplified soft wrist composed of four degrees of freedom: $z$, roll, pitch, and yaw. The stiffness parameters were the same as those of the previous study~\cite{fuchioka2024robotic}. 
We assume that the peg was fixed to the gripper and no slippage occurred during the tasks. 
The tolerance between the peg and hole was 2 mm.
We used a green and red colored peg and hole and applied OpenCV for pose estimation from vision, introduced in Sec.~\ref{sec:multimodal_verification}.

{\bf Default condition and initial goal parameters:} We designed the initial goal pose $p_g$ in the default condition where the circular peg was aligned in the gripper, and the hole was fixed in a specific position. We controlled the robot with a keyboard and recorded $p^n_g$ in $n$ contact formations.

{\bf Randomized conditions:} To see the robustness of the failure recovery, we tested our method in uncertain environments. 
We uniformly randomized the in-grasp angle of the peg within $\pm 5${\degree}, the uncertainty of the position of the hole within $\pm 20$ mm from the default poses. The hole's surface friction was randomized up to a 5-times increase compared to the default condition. 
We also tested our method with unseen peg shapes, including squares and rectangles.

{\bf Metrics and baselines:} The robot sequentially executed skills given the default goal pose $p_g$ and our recovery method in the four conditions, including (1) no randomization, (2) peg and hole pose randomization, (3) friction randomization, and (4) all randomization (peg and hole pose and friction). 
As a baseline, we executed the skills with the default goal pose and without our recovery method.

The trials were performed 30 times, and the success rates of the overall task completions were evaluated. 
Overall success was defined as when the peg was inserted into the hole to a depth of 10 mm.
Overall failures were considered when a timeout (i.e., recovery skill executions failed consecutively 10 times) occurred.

{\bf Results:} The success rates for each condition are shown in Table~\ref{tab:sim_success_rate}. The values in each contact formation (Fit, Align, and Insertion) denote the average number of additional executions triggered by the failure recovery.  The baseline without recovery achieves 47\% success under the all randomization setting. In contrast, our recovery-enabled method improves the success rate to 83\%, underscoring the significant contribution of recovery. In addition, the results show that the proposed framework generalizes robustly to previously unseen peg shapes (square and rectangular); notably, it achieves 80\% success under the all randomization setting with the square peg.

% sim transition matrix
The recovery transitions of the circular peg are presented in Table~\ref{tab:sim_transition_matrix}. Although the majority of the transitions are the same contact formations, it should be noted that among the recovery strategies called upon failures in Align execution, 9.5\% of them were Fit.
%\vspace{0.5em}

% \usepackage{tabularray}
\begin{table}
%\vspace{-1em}
\centering
\caption{Performance under different conditions for each peg shape in simulation. ``F" denotes framework success (i.e., the VLM’s judgment of task completion), while ``G" denotes ground-truth success/failure.}
\label{tab:sim_success_rate}
\begin{tblr}{
  row{1} = {c},
  cell{1}{1} = {r=2}{},
  cell{1}{2} = {r=2}{},
  cell{1}{6} = {r=2}{},
  cell{2}{3} = {c=3}{},
  cell{3}{1} = {r=4}{c},
  cell{3}{3} = {c},
  cell{3}{4} = {c},
  cell{3}{5} = {c},
  cell{3}{6} = {c},
  cell{4}{3} = {c},
  cell{4}{4} = {c},
  cell{4}{5} = {c},
  cell{4}{6} = {c},
  cell{5}{3} = {c},
  cell{5}{4} = {c},
  cell{5}{5} = {c},
  cell{5}{6} = {c},
  cell{6}{3} = {c},
  cell{6}{4} = {c},
  cell{6}{5} = {c},
  cell{6}{6} = {c},
  cell{7}{1} = {r=4}{c},
  cell{7}{3} = {c},
  cell{7}{4} = {c},
  cell{7}{5} = {c},
  cell{7}{6} = {c},
  cell{8}{3} = {c},
  cell{8}{4} = {c},
  cell{8}{5} = {c},
  cell{8}{6} = {c},
  cell{9}{3} = {c},
  cell{9}{4} = {c},
  cell{9}{5} = {c},
  cell{9}{6} = {c},
  cell{10}{3} = {c},
  cell{10}{4} = {c},
  cell{10}{5} = {c},
  cell{10}{6} = {c},
  cell{11}{1} = {r=4}{c},
  cell{11}{3} = {c},
  cell{11}{4} = {c},
  cell{11}{5} = {c},
  cell{11}{6} = {c},
  cell{12}{3} = {c},
  cell{12}{4} = {c},
  cell{12}{5} = {c},
  cell{12}{6} = {c},
  cell{13}{3} = {c},
  cell{13}{4} = {c},
  cell{13}{5} = {c},
  cell{13}{6} = {c},
  cell{14}{3} = {c},
  cell{14}{4} = {c},
  cell{14}{5} = {c},
  cell{14}{6} = {c},
  cell{15}{1} = {r=4}{c},
  cell{15}{3} = {c},
  cell{15}{4} = {c},
  cell{15}{5} = {c},
  cell{15}{6} = {c},
  cell{16}{3} = {c},
  cell{16}{4} = {c},
  cell{16}{5} = {c},
  cell{16}{6} = {c},
  cell{17}{3} = {c},
  cell{17}{4} = {c},
  cell{17}{5} = {c},
  cell{17}{6} = {c},
  cell{18}{3} = {c},
  cell{18}{4} = {c},
  cell{18}{5} = {c},
  cell{18}{6} = {c},
  hline{1,19} = {-}{0.08em},
  hline{2} = {3-5}{},
  hline{3} = {-}{},
  hline{4-6,8-10,12-14,16-18} = {2-6}{dotted},
  hline{7,11,15} = {-}{0.05em},
}
{Peg \\Shape} & {Randomized \\Conditions} & Fit & Align & Insert & {Succ. \\{[}F, G]}\\
 &  & Avg. \# extra executions &  &  & \\
\shortstack{$\bigcirc$\\Baseline \\(w/o recovery)} & None & — & — & — & 100, 100\\
 & Friction & — & — & — & 43, 43\\
 & {Hole Pose \&\\Peg Angle} & — & — & — & 57, 57\\
 & All & — & — & — & 47, 47\\
$\bigcirc$ & None & 0.0 & 0.0 & 0.0 & 100, 100\\
 & Friction & 0.6 & 0.1 & 0.0 & 100, 100\\
 & {Hole Pose \&\\Peg Angle} & 0.7 & 1.1 & 0.7 & 93, 90\\
 & All & 0.8 & 1.0 & 1.1 & 87, 83\\
\shortstack{{$\Box$} \\ (Unseen)} & None & 0.0 & 0.0 & 0.0 & 100, 100\\
 & Friction & 0.0 & 0.3 & 0.0 & 100, 100\\
 & {Hole Pose \&\\Peg Angle} & 0.4 & 1.3 & 0.5 & 93, 93\\
 & All & 0.8 & 2.1 & 1.1 & 80, 80\\
\shortstack{$\rectangle$ \\ (Unseen)} & None & 0.0 & 0.0 & 0.0 & 100, 100\\
 & Friction & 0.0 & 0.0 & 0.0 & 100, 100\\
 & {Hole Pose \&\\Peg Angle} & 0.7 & 2.1 & 0.8 & 77, 77\\
 & All & 0.6 & 2.5 & 1.5 & 67, 67
\end{tblr}
\end{table}

\begin{table}
% \vspace{-1em}
\centering
\caption{Transition Matrix of Circular Peg in simulation}
\label{tab:sim_transition_matrix}
\begin{tabular}{|c|c|c|c|}
\hline
\textbf{From\textbackslash To} & \textbf{Fit} & \textbf{Align} & \textbf{Insert} \\
\hline
\textbf{Fit}    & 100.0\% & 0.0\%   & 0.0\% \\
\hline
\textbf{Align}  & 9.5\%   & 90.5\%  & 0.0\% \\
\hline
\textbf{Insert} & 0.0\%   & 1.8\%   & 98.2\% \\
\hline
\end{tabular}
\end{table}
%\vspace{-0.8em}

\subsection{Real robot experiments}
To answer the third question, we evaluated the proposed recovery framework on a real robot using a round-shaped peg. We used an Intel RealSense D435f camera. Experiments were conducted under seven combinations of conditions, varying peg tilt angles (0° and 5°) and hole positions (0, 10, 15, and 20 mm). The success rates in five trials for each condition are shown in Table~\ref{tab:real_success_rate}. The results confirmed that, with recovery, a success rate of 80\% or higher was achieved in all conditions. Similar to the simulation, the number of additional executions per subtask increased, indicating that the condition change necessitated error recovery for task completion. Fig.~\ref{fig:snapshots} shows snapshots of our recovery.

\begin{table}
%\vspace{-1em}
\centering
\caption{Performance under different combinations of hole position and peg angle in real robot experiments}
\label{tab:real_success_rate}
\begin{tabular}{cccccc} 
\toprule
\multirow{2}{*}{Hole} & \multirow{2}{*}{Angle} & Fit & Align & Insert & \multirow{2}{*}{Success [F, G]} \\ 
\cmidrule{3-5}
 &  & \multicolumn{3}{l}{Avg. \# extra executions} &  \\ 
\hline
0 & 0 & 0.0 & 0.2 & 0.6 & 100, 100 \\
10 & 0 & 0.2 & 0.4 & 0.2 & 100, 100 \\
15 & 0 & 1.4 & 1.4 & 0.4 & 100, 100 \\
20 & 0 & 2.2 & 2.8 & 0.0 & 80, 80 \\
10 & 5 & 2.0 & 0.8 & 0.2 & 100, 100 \\
15 & 5 & 2.0 & 0.4 & 0.2 & 100, 100 \\
20 & 5 & 3.0 & 1.8 & 1.4 & 80, 100 \\
\bottomrule
\end{tabular}
\end{table}

%% file: sections/discussion.tex
\subsection{Discussion about experimental results}
{\bf All randomization settings with a circular peg:}
In the all randomization condition (hole pose, peg angle, and friction) with a circular peg, the success rate was 83\%. Log analysis suggests that the main factor limiting further improvement was activation of the wrist’s compression limit during repeated Insertion attempts. This occurred when small misalignments of the peg persisted, and the Align success check, affected by vision estimation error, classified the alignment as successful. After one or more retries, the system reached the predefined maximum number of trials (set to 10 in this study).

{\bf Accuracy of success classification:} 
The framework’s classifications (F) were generally consistent with the ground truth (G), reflecting the effect of the verification mechanisms. This indicates that the method provides reliable success judgments across diverse and randomized conditions. Occasional discrepancies were observed; for instance, the framework classified a task as successful when the peg was not fully inserted.

{\bf Robustness against friction variation:} Although a previous study~\cite{Morgan2023TowardsGeneralizedRobotAsembly} reported that compliance-enabled contact formation has robustness to variations in hole pose and peg angle, it struggled with higher friction conditions. 
In contrast, the proposed recovery framework achieved a 100\% success rate under the increased friction condition. 

\textbf{Alignment failures with unseen peg shapes: }
Under the all-randomization setting, the success rate declined by 7\% for the square peg and 20\% for the rectangular peg compared with the circular peg, as shown in Table~\ref{tab:sim_success_rate}.
The primary factor for this decline was the increased likelihood of the robot reaching the maximum number of predefined recovery attempts in the Align subtask. 

For example, under the \textit{Hole Pose \& Peg Angle} condition, 
the average number of Align executions was 1.1 for the circular peg, 1.3 for the square peg, and 2.1 for the rectangular peg.
% the average number of executions for Align increased by 0.2 times for the square peg and 0.8 times for the rectangular peg compared to the circular peg. 
Log analysis revealed that failures predominantly occurred when the randomized angle exceeded 4.0$\degree$. In many of these failures, as the system attempted to recover from misalignment, the peg frequently overshot the hole, attempted to correct, and overshot again. The challenge was to fit the peg into the hole such that, by leveraging the compliance of the wrist, the peg could be aligned to an angle close to 0$\degree$. 
When the initial angle was large, this became increasingly difficult, as the vertical height of the peg shortened with tilt, making proper fitting more challenging.

\textbf{Real robot experiments:} 
Real robot experiments achieved a high success rate, with one failure observed at hole = 20 mm and angle = 0$\degree$, where repeated misalignments during the Align subtask caused the system to reach the maximum number of retries.
Analysis revealed that this was caused by an excessively large parameter update suggested by the VLM. This suggests that such failures could be mitigated by narrowing the allowable range for parameter adjustments at each step.

\subsection{Limitations and future work}
Our method demonstrated successful recoveries in randomized and unseen conditions; however, some limitations remain.
First, we employed a simple pose
estimation method using OpenCV for pegs and holes of a specific color and shape. Open vocabulary object estimation
methods~\cite{corsetti2024open} can be used for pegs of diverse colors and shapes.
Second, this study used pre-designed skill sets. They can be replaced with more intelligent learning-based policies that may reduce the number of recovery retries. Nevertheless, even with more intelligent policies, some failures are likely to occur. Thus, our method will remain necessary across different skill sets.
For future work, we will consider integrating learning-based skills described above into our compliance-enabled contact formation and failure recovery framework.